\documentclass[10pt,twocolumn,letterpaper]{article}

\usepackage{cvpr}              
\definecolor{cvprblue}{rgb}{0.21,0.49,0.74}
\usepackage[pagebackref,breaklinks,colorlinks,allcolors=cvprblue]{hyperref}

\usepackage{multirow}
\usepackage{pifont}
\newcommand{\cmark}{\ding{51}}%
\newcommand{\xmark}{\ding{55}}%

\def\paperID{6} 
\def\confName{CVPRW}
\def\confYear{2026}

\title{Customized Visual Storytelling with Unified Multimodal LLMs}

\author{Wei-Hua Li\textsuperscript{1}, Cheng Sun\textsuperscript{2}, Chu-Song Chen\textsuperscript{1} \\
\textsuperscript{1}National Taiwan University, \textsuperscript{2}NVIDA\\
{\tt\small \{d12922009,chusong\}@csie.ntu.edu.tw,} {\tt\small chengs@nvidia.com}\\
}

\begin{document}
\maketitle

\begin{abstract}
Multimodal story customization aims to generate coherent story flows conditioned on textual descriptions, reference identity images, and shot types. While recent progress in story generation has shown promising results, most approaches rely on text-only inputs. A few studies incorporate character identity cues (e.g., facial ID), but lack broader multimodal conditioning.
In this work, we introduce VstoryGen, a multimodal framework that integrates descriptions with character and background references to enable customizable story generation. To enhance cinematic diversity, we introduce shot-type control via parameter-efficient prompt tuning on movie data, enabling the model to generate sequences that more faithfully reflect cinematic grammar.
To evaluate our framework, we establish two new benchmarks that assess multimodal story customization from the perspectives of character and scene consistency, text–visual alignment, and shot-type control. Experiments demonstrate that VstoryGen achieves improved consistency and cinematic diversity compared to existing methods.
\end{abstract}

\section{Introduction}
\label{sec:intro}

Recent advancements in text-to-video (T2V) or text-and-image-to-video (TI2V)~\cite{kondratyuk2023videopoet, liu2025video, wan2025wan, deepmind2025veo3} transformers have shown notable progress in generating compelling videos; however, these outputs remain limited to short clips. Several studies~\cite{chen2023seine, wang2025lingen, dalal2025one} have extended video durations, but the resulting videos are often restricted to slow motion. Because transformer computation and memory costs increase quadratically, sustaining long-term generation is highly demanding. To address this, research~\cite{he2025dreamstory} has introduced pipelines that extend video length by first using text-to-image (T2I) models to generate \textit{keyframes}, which are then expanded into video clips via image-to-video (I2V) or TI2V models. Concatenating these clips enables longer video generation, with keyframes serving as structural anchors for storytelling. Keyframe-based methods align more naturally with storylines and scale effectively to long-form video creation, yet producing videos with structured narratives and coherent characters and scenes remains a challenge.

Another related line of research, visual story generation, aims to produce a series of images based on text inputs~\cite{tewel2024training, zhou2024storydiffusion, wang2025characonsist, liu2025one, ma2025storynizor}. These methods often rely on T2I diffusion models such as stable or latent diffusion~\cite{rombach2022high}, diffusion transformers~\cite{peebles2023scalable}, or FLUX.1~\cite{labs2025flux}, with CLIP~\cite{radford2021learning} or related models used to encode the input text. Some methods~\cite{liu2025one, ma2025storynizor, zhou2024storydiffusion, tewel2024training} can maintain the consistency of particular subjects (e.g., character faces) generated from text prompts.
{\color{black} However, background consistency is also crucial for story generation to ensure narrative coherence.} CharaConsist~\cite{wang2025characonsist}, therefore, improves scene consistency derived from input text. {\color{black} Despite these advances, they cannot leverage reference images as inputs for guiding character- or scene-specific customization.} Moreover, their ability to comprehend complex textual descriptions remains weak, hindering the generation of detailed visual outputs.

\begin{figure}[!t]
    \centering
        \includegraphics[width=1\columnwidth]{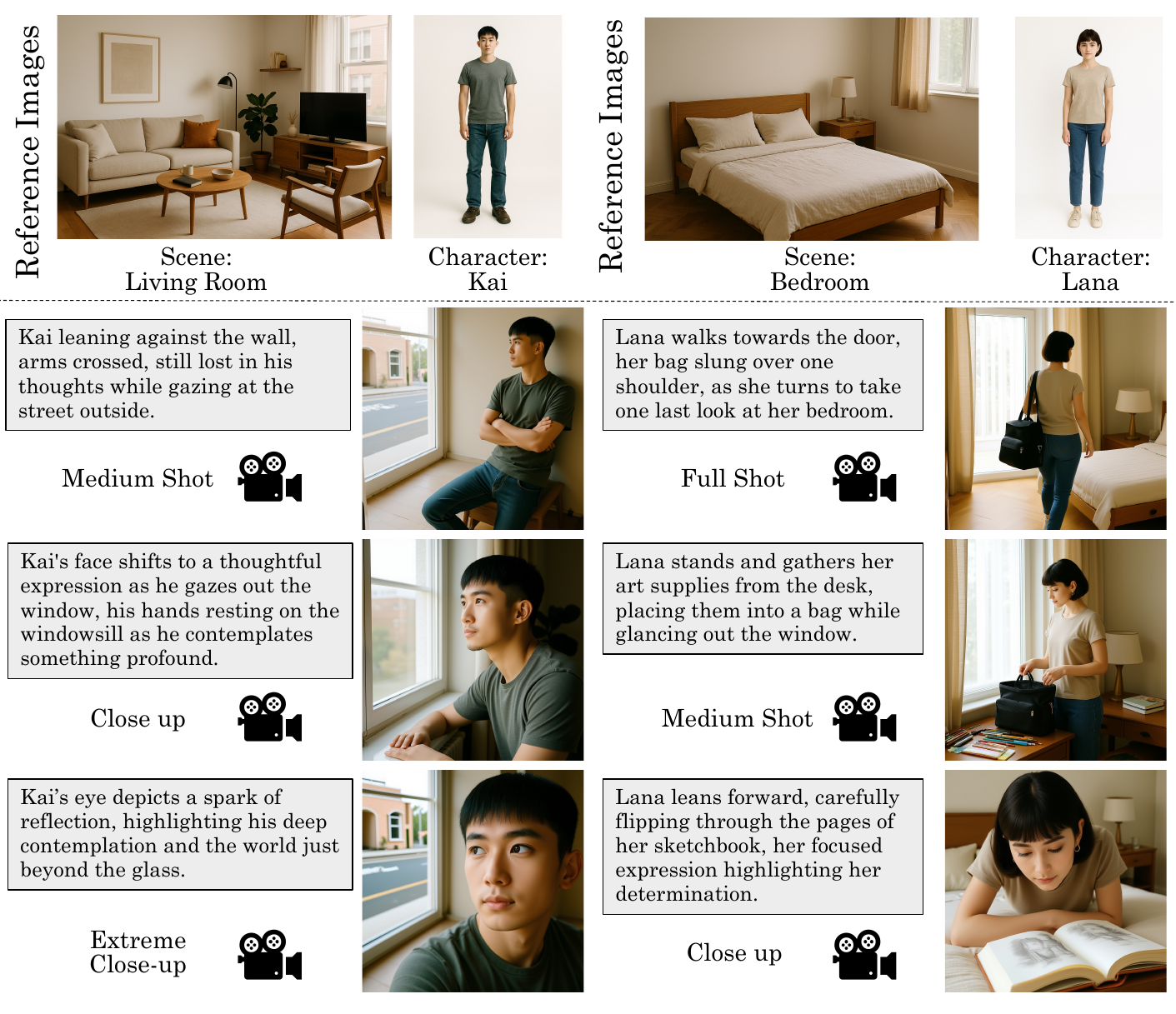}
        \caption{Shot-Type Controlled Visual Story Customization}
        \label{fig:teaser}
        \vspace{-10pt}
\end{figure}


To address the aforementioned problems, we introduce \textbf{VstoryGen} for multimodal story customization, which conditions on textual descriptions, reference images, and shot-type controls to generate coherent story sequences. 
Our approach is inspired by the customization concept in image editing~\cite{kong2025custany, ye2023ip}. Reference images and their associated textual descriptions help customize the story sequences with temporal consistency. In addition, customization guidance enables the produced videos to better align with the structured narrative of storytelling. To handle multimodal inputs, the capability to comprehend intricate textual descriptions and their association with images is essential. We leverage the remarkable ability of multimodal large language models (MLLMs) to understand both text and images. In particular, we employ unified MLLMs (UMLLMs)~\cite{chen2025janus, xiao2025omnigen}, as they can generate multimodal outputs. 



VstoryGen consists of three stages that transform free-form descriptions into video sequences, as illustrated in Fig.~\ref{fig:vstorygen}. First, \textbf{multimodal script generation}, a preprocessing step that constructs structured multimodal scripts (including text prompts, reference images, and shot types) to guide keyframe production. Second, \textbf{CustFilmer} generates {\color{black} an arbitrary number of} customized and consistent keyframes according to the scripts.
Finally, \textbf{video generation}, which takes the keyframes as anchors and leverages existing TI2V models to expand them into videos. 
In addition, CustFilmer incorporates shot-type control through prompt tuning with movie data from CMD~\cite{bain2020condensed}. This design enables CustFilmer to produce more cinematic views, overcoming the monotony of fixed-view generation in existing methods, as shown in Fig~\ref{fig:teaser}. 


For evaluation, existing benchmarks (e.g., DS500~\cite{tewel2024training}, ConsiStory+~\cite{liu2025one}) are limited to text-only inputs. These datasets do not suit assessing the performance of story customization. To properly evaluate our method, we introduce the Multimodal Storyboard Benchmark (MSB) and the Multimodal \& Multisubject Storyboard Benchmark (M$^2$SB), designed to assess multimodal story customization in single- and multi-subject settings, respectively. Both benchmarks include diverse identity images, backgrounds, narrative scripts, and shot-type annotations, providing a unified basis for fair comparison.


Our contributions are summarized as follows:
\begin{itemize}
\item We propose \textbf{VstoryGen}, a pipeline for long-form story generation with multimodal customization. It provides a framework to support customization based on text prompts, character(s)/scene references, and shot types.
\item We release two new benchmarks, \textbf{MSB} and \textbf{M$^{\mathbf{2}}$SB}, which contain multimodal materials and provide standardized evaluation for multimodal story customization.
\end{itemize}

\section{Related Works}
\label{sec:related}
In this section, we review several related research topics, including long-form video generation models, MLLMs \& UMLLMs, and visual story generation.


\subsection{Long-form Video Generation Models}\label{sec:long}
Long-form video generation has been a long-standing challenge in the field. Recent works~\cite{chen2023seine, wang2025lingen, dalal2025one} have extended video durations to 
about one minute, surpassing the lengths achievable by models such as Sora~\cite{brooks2024video} and Veo3~\cite{deepmind2025veo3}. SEINE~\cite{chen2023seine} is designed to reuse the last few frames of a generated video to predict subsequent ones. LingGen~\cite{wang2025lingen} enables linear-complexity generation but remains limited to single scenes and slow motion, without capturing complex narratives. TTT-Video~\cite{dalal2025one} produces clips up to one minute long. To further extend video length, existing methods~\cite{he2025dreamstory, he2025dreamstory} adopt a multi-stage pipeline that first generates keyframes, expands them into short clips via I2V models, and concatenates them into long videos. However, these methods primarily focus on foreground consistency while neglecting scene and contextual coherence, resulting in inconsistent scenes and reduced expressiveness.

\subsection{MLLMs and UMLLMs} 
In recent years, MLLMs~\cite{liu2023visual, bai2025qwen2} have extended the capabilities of LLMs~\cite{touvron2023llama} by projecting visual features and aligning them with textual representation, thereby enabling strong image understanding and reasoning ability.
However, MLLMs can only output text.
At the same time, image generation with diffusion-based methods~\cite{rombach2022high, gu2022vector, peebles2023scalable} and autoregressive-based methods~\cite{esser2021taming} have demonstrated powerful generative capacity. Building on these advances in both multimodal understanding and generative model, recent efforts have explored UMLLMs~\cite{zhang2025unified} for multimodal understanding and generation. 
{Existing UMLLMs can be divided into three categories based on their backbone architecture: (1) Diffusion Models, (2) Autoregressive models and (3) Fused AR and Diffusion Models.

\subsubsection{Diffusion Models} 
Diffusion models have demonstrated remarkable success in image generation. Building on these advances, recent studies have extended diffusion-based frameworks from unimodal settings to multimodal generation, enabling unified modeling of both textual and visual outputs. For instance, Dual Diffusion~\cite{Li_2025_CVPR} jointly models image and text distributions through a unified denoising diffusion objective. MMaDA~\cite{yang2025mmada}, a multimodal diffusion foundation model, further introduces a mixed chain-of-thought fine-tuning strategy to unify reasoning formats across vision and language tasks. Despite their promising capabilities, a major limitation of diffusion-based multimodal models lies in inference efficiency. 

\subsubsection{Autoregressive (AR) Models}
AR models have emerged as the dominant paradigm in existing UMLLMs. These models are typically built upon Transformer backbones adapted from LLMs, such as the LLaMA series~\cite{touvron2023llama} and the Qwen series~\cite{bai2025qwen2, yang2025qwen3}. To incorporate visual information, prior studies explore diverse image tokenization strategies during encoding, including semantic tokenization with Vision Transformer (ViT)~\cite{dosovitskiy2020image} and pixel-level tokenization with Variation Autoencoder (VAE)~\cite{kingma2013auto}. 

Within this unified AR framework, multimodal embeddings are autoregressively predicted and subsequently decoded into text via detokenization modules or visual outputs via  pixel-based decoders (e.g., VQVAE~\cite{razavi2019generating}, VQGAN~\cite{esser2021taming}) or diffusion-based decoders (e.g., Diffusion Transformer (DiT)~\cite{peebles2023scalable}). Seed-X~\cite{ge2024seed} introduces multi-granularity modeling to support image understanding and generation across arbitrary resolutions. Emu3~\cite{wang2024emu3} adopts a fully autoregressive formulation by tokenizing all modalities into a single sequence, enabling unified image and video generation. Janus-Pro~\cite{chen2025janus} employs decoupled image encoders to separate semantic understanding from visual synthesis. OmniGen2~\cite{wu2025omnigen2} further integrates text-to-image synthesis, image editing, and in-context generation within a rectified-flow framework. These AR-based UMLLMs demonstrate the strong visual understanding and image synthetic ability.

\subsubsection{Fused AR and Diffusion Models}
Fused AR and diffusion modeling has recently emerged as a powerful paradigm for unified vision–language generation. In this framework, textual tokens are produced in an autoregressive manner, inheriting strong compositional reasoning capabilities of LLMs, while visual content is synthesized through iterative denoising processes following the diffusion principle. By combining the complementary strengths of these two paradigms, fused AR–diffusion models achieve both flexible language generation and high-quality image synthesis within a unified architecture. 

Several recent works exemplify this hybrid design. Show-o~\cite{xie2024show} integrates autoregressive text modeling with diffusion-based image generation to support multimodal content creation. Transfusion~\cite{zhou2024transfusion} further explores cross-modal token alignment between AR and diffusion branches, enabling coherent joint generation of text and images. Janus-Flow~\cite{ma2025janusflow} adopts a rectified-flow formulation to unify autoregressive language modeling with diffusion-based visual synthesis, demonstrating improved stability and generation quality across multimodal tasks.

In this paper, we focus on AR models for story customization, as they represent the dominant direction in UMLLMs. {\color{black}Moreover, AR models naturally support generating story sequences of arbitrary length, enabling flexible and scalable story generation.}

\subsection{Visual Story Generation}
The goal of visual story generation is to produce coherent sequences of images that ensure both character and scene consistency while preserving logical continuity throughout the narrative. Prior works have explored this challenge from different perspectives, with early studies focusing primarily on character (foreground) consistency. 

For instance, ConsiStory~\cite{tewel2024training} is designed to maintain subject identity while aligning with textual descriptions; StoryDiffusion~\cite{zhou2024storydiffusion} incorporates Consistent Self-Attention to enhance character preservation; and Storynizor~\cite{ma2025storynizor} introduces ID-Injector and ID-Synchronizer modules for identity consistency. Similarly, StoryWeaver~\cite{zhang2025storyweaver}, DreamStory~\cite{he2025dreamstory}, and 1P1S~\cite{liu2025one} continue to prioritize character consistency. Although these methods improve character preservation, they remain limited in enforcing scene coherence, often resulting in discontinuities. CharaConsist~\cite{wang2025characonsist} advances this direction by leveraging point-tracking attention and adaptive token merging to align foreground and background, but it still relies solely on text inputs, restricting effective customization. 

In contrast, our approach conditions on a comprehensive set of multimodal materials, including text prompts, foreground subjects, background scenes, and shot-type annotations. This provides greater flexibility for user customization, whereas prior approaches cannot accommodate such diverse conditions.

\begin{figure}[!t]
    \centering
    \includegraphics[width=1\columnwidth]{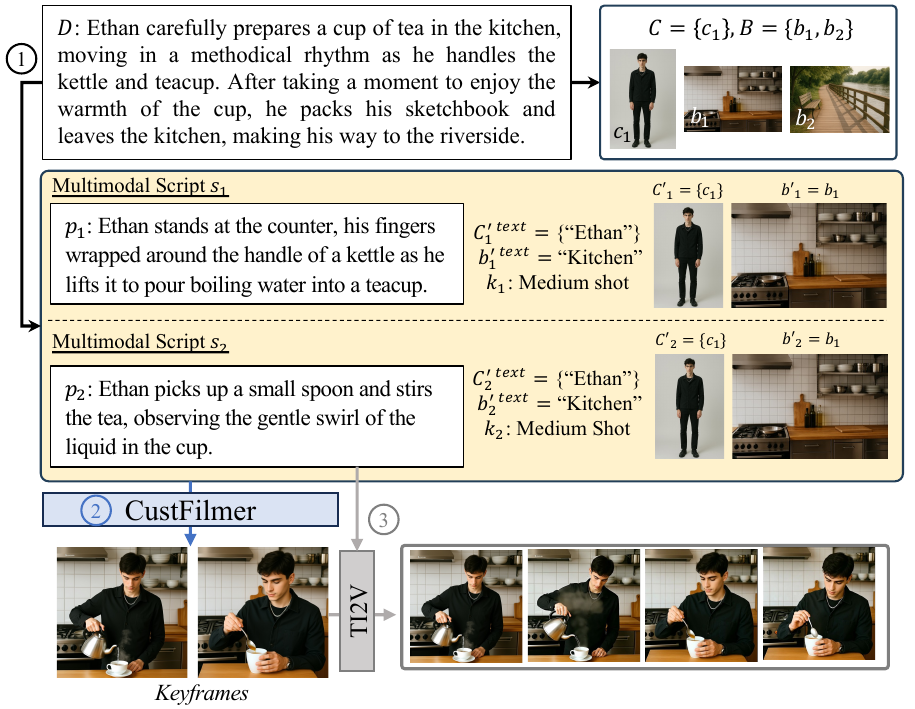}
    \caption{Overview of VstoryGen. (1) Multimodal scripts are generated from an overall text description $D$, where the reference images of characters (indexed $1$ to $m$) and background scenes (indexed $1$ to $o$) are generated first, and the multimodal scripts (indexed $1$ to $n$) are generated afterward; (2) CustFilmer produces consistent keyframes corresponding to the scripts from $1, \cdots, n$, respectively; (3) TI2V expands the $n$ keyframes into video.}
    \label{fig:vstorygen}
\end{figure}

\begin{figure*}[!t]
    \centering
    \includegraphics[width=\linewidth]{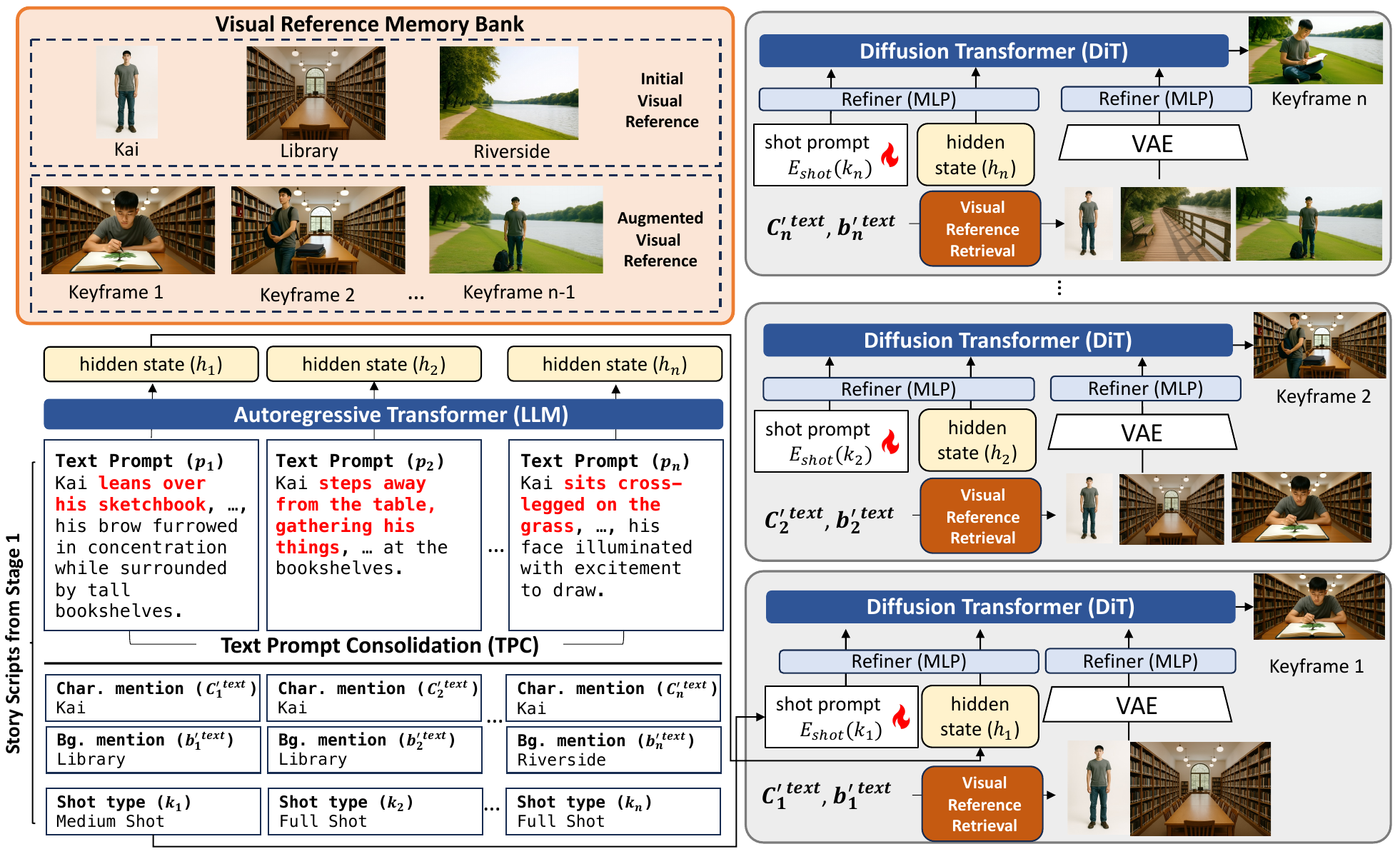}
    \caption{Illustration of CustFilmer, which can take multimodal materials as input to generate sequences of consistent keyframes.}
    \label{fig:custfilmer}
    \vspace{-5pt}
\end{figure*}

\section{Method}
\label{sec:method}





VstoryGen generates coherent sequences with consistent characters, scenes, and diverse content.
The solution is scalable to long video generation. 
First, Multimodal Script Generation uses
a \textbf{Multimodal Generator as Director} to structure the narrative into multimodal scripts. 
Second, \textbf{CustFilmer} customizes the keyframes based on the multimodal scripts. 
We then leverage existing TI2V models to expand generated keyframes into videos.

\subsection{Multimodal Generative Model as Director} 
Existing T2I story generation methods~\cite{he2025dreamstory, wang2025characonsist} commonly rely on powerful generative models (e.g., GPT-4) to produce a sequence of text prompts from a story description. Following this practice, we leverage GPT-4o to prepare text prompts and shot types for each keyframe, as well as reference images (character and background) for the story, and construct the initial visual reference for the next step. For each keyframe, we construct a script $s_t$ ($t = 1, \ldots, n$) based on a free-form story description $D$,  as illustrated in Fig.~\ref{fig:vstorygen}.

Given a description $D$, GPT-4o produces text prompts 
$P = \{p_1, \ldots, p_n\}$, each describing a corresponding keyframe. 
%
The script $s_t = (\, p_t,\;  C_t^{'\text{text}},\; b_t^{'\text{text}},\; k_t \,)$ consists of a textual prompt $p_t$, selected textual mentions of character(s) $C_t^{'\text{text}} = \{c_{t,0}^{'text} ... c_{t,q}^{'text}\} $ and background(s) $b_t^{'\text{text}}$, 
and a shot type $k_t \in \mathcal{K}$, where $q$ denotes the number of characters in each script and $\mathcal{K}$ denotes the shot-type vocabulary set.
The scripts serve as conditions for the next stage, where CustFilmer instantiates them into consistent keyframes.




\subsection{CustFilmer as a Keyframes Generator}
\subsubsection{AR-based UMLLM for Image Editing} 
{One major line of research in UMLLMs adopts AR-based architectures, where multimodal tokens are serialized and modeled sequentially within a unified Transformer backbone. Recent AR-based UMLLMs further extend this paradigm to image and video editing~\cite{wu2025qwen, xiao2025omnigen, wu2025omnigen2, wei2025univideo}. Instead of injecting visual features into the AR backbone, these approaches encode reference images or videos with a VAE and deliver the embeddings directly to the diffusion decoder, thereby better preserving low-level visual information.
While such designs significantly improve editing fidelity, they remain primarily focused on single-step content modification and are not optimized for generating long sequences of coherent story frames. 
Building upon this AR-based editing UMLLM paradigm, we propose CustFilmer, enabling scalable and temporally consistent multi-image generation within a unified model.

\subsubsection{CustFilmer}
Given a multimodal script from the previous stage, the goal of CustFilmer is to generate the corresponding keyframes. 
A straightforward extension is to adopt multi-turn interactions, enabling UMLLMs to generate multiple images sequentially. However, such multi-turn generation is highly inefficient and prone to error accumulation across turns, making it difficult to preserve long-term consistency. 

To address these limitations, we propose CustFilmer that restructures existing UMLLMs into an autoregressive model with iterative diffusion-based decoders (e.g. DiT~\cite{peebles2023scalable}), enabling scalable story image generation. In CustFilmer, we introduce three mechanisms to better adapt UMLLMs for visual storytelling: (1) \textit{Text Prompt Consolidation}, (2) \textit{Keyframe-wise Autoregressive Generation} and (3) \textit{Visual Reference Memory Bank and Retrieval}. These mechanisms improve intra-frame consistency while preserving the base model’s customization and generation flexibility. An illustration is shown in Fig.~\ref{fig:custfilmer}




\paragraph{Text Prompt Consolidation (TPC).} 
To achieve consistent T2I generation, prior work~\cite{liu2025one} shows that language models exhibit context consistency, enabling identity and semantic information to be preserved within a consolidated prompt, resulting in coherent CLIP embeddings~\cite{radford2021learning}.

Inspired by this property, we propose TPC, which jointly encodes prompts using LLMs to produce text embeddings with stronger semantic and identity consistency than independent encoding. Specifically, for each story, we group their text prompts $P = \{p_1, \dots, p_n\}$ in the same batch for model inference. We then extract their hidden states $H = \{h_1, \dots, h_n\}$ through an auto-regressive model, as illustrated in Fig.~\ref{fig:custfilmer}. These hidden states represent different events within the same story and serve as semantic conditions for keyframe generation in the subsequent stage. 

\paragraph{Keyframe-wise Autoregressive Generation.} 
%
To enable the generation of arbitrarily long image sequences, we propose a keyframe-wise autoregressive generation framework that propagates temporal information across keyframes.
Following prior AR-based UMLLMs for image editing, we employ a VAE to encode pixel-level features from reference images (e.g., character portraits and background scenes) and inject them into the diffusion decoder (e.g., DiT). 
These reference images are dynamically retrieved based on the current context, including both predefined inputs and previously generated keyframes, as detailed in the next section.
At each step $t$, the DiT module synthesizes the current keyframe as 
$I_t = \mathrm{DiT}(h_t, z_t)$, where $h_t$ denotes the hidden state produced by the AR model, and $z_t$ denotes the VAE-encoded visual features of the reference images.

\paragraph{Visual Reference Memory Bank and Retrieval.}
{To enhance temporal consistency while avoiding reference ambiguity, we introduce a Visual Reference Memory Bank with a structured retrieval.} 
The memory bank stores both initial reference images (from Stage 1) and previously generated keyframes, organized as a key–value dictionary. Specifically, at each timestep $t$, the character mentions $C^{\text{text}}_t$ and background mentions $b^{\text{text}}_t$ serve as queries to retrieve the corresponding visual references. 
Unlike embedding-based retrieval, our approach leverages structured script annotations, ensuring precise and interpretable reference selection. 

In addition to static references, we further retrieve the most recent $\mu$ previously generated keyframes $\{I_{t-i}\}_{i=1}^{\mu}$ that share the same character and scene, providing temporally coherent visual cues. 
This design jointly enforces semantic alignment (via structured queries) and temporal consistency (via recent frames).

The retrieved references are then aggregated and encoded as:
$ z_t = \mathrm{VAE}\big[\mathcal{R}_t, \{\textit{Scale}_\alpha(I_{t-i})\}_{i=1}^{\mu}\big]$, where $\mathcal{R}_t = \{C'_t, b'_t\}$ denotes the retrieved character and background references, and $\alpha$ controls the trade-off between consistency and diversity. By combining memory retrieval with recursive conditioning on previous frames, our design enables coherent visual information propagation across keyframes.

\begin{figure*}[!ht]
    \centering
    \includegraphics[width=.95\linewidth]{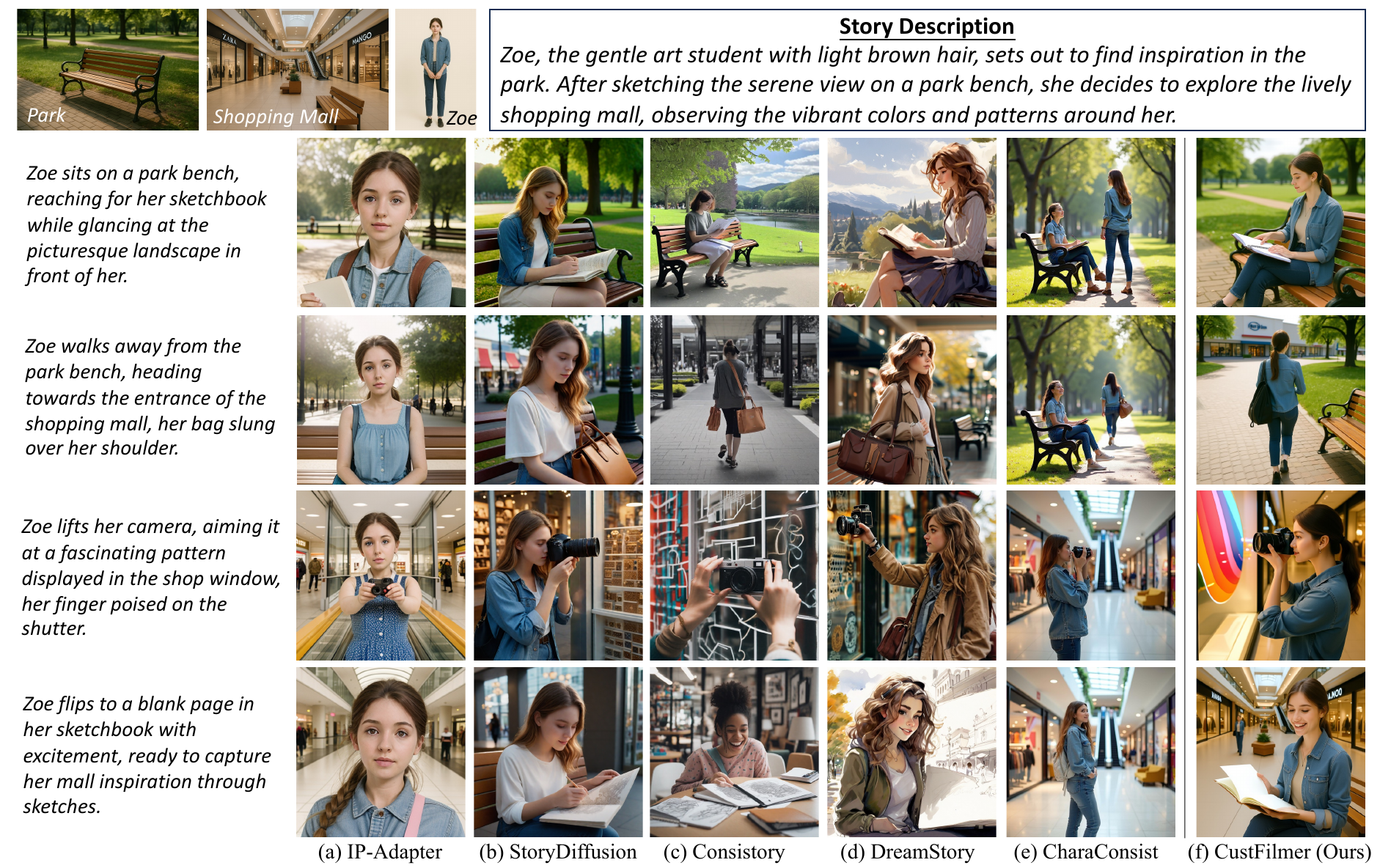}
    \caption{Qualitative Comparison between (a) IP-Adapter~\cite{ye2023ip}, (b) StoryDiffusion~\cite{zhou2024storydiffusion}, (c) Consistory~\cite{tewel2024training}, (d) DreamStory~\cite{he2025dreamstory}, (e) CharaConsist~\cite{wang2025characonsist} and (f) CustFilmer (Ours) on MSB}
    \label{fig:qc1}
    \vspace{-5pt}
\end{figure*}

\begin{table*}[!t]
\centering
\caption{Quantitative comparison of prior methods on consistency metrics.}
\resizebox{.95\textwidth}{!}{
\begin{tabular}{llccccc}
\toprule
\multirow{2}{*}{Method} & \multirow{2}{*}{Base Model} 
  & \multicolumn{2}{c}{Inter-Consistency} 
  & \multicolumn{2}{c}{Intra-Consistency} 
  & \multirow{2}{*}{\shortstack{Average\\Consistency ($\uparrow$)}} \\
\cmidrule(lr){3-4} \cmidrule(lr){5-6}
 &  & CLIP-I-fg($\uparrow$) & CLIP-I-bg($\uparrow$) 
    & CLIP-I-fg($\uparrow$) & CLIP-I-bg($\uparrow$) &  \\
\midrule
IP-Adapter (Arxiv'23)     & \textit{SDXL} & 0.901 & 0.936 & 0.900 & 0.646 & 0.846 \\
Consistory (SIGGRAPH'24)  & \textit{SDXL} & 0.868 & 0.884 & 0.883 & 0.612 & 0.812 \\
StoryDiffusion (NeurIPS'24) & \textit{SDXL} & 0.857 & 0.900 & \textbf{0.921} & 0.645 & 0.831 \\
DreamStory (TPAMI'25)     & \textit{SDXL} & 0.844 & 0.858 & 0.896 & 0.635 & 0.808 \\
CharaConsist (ICCV'25)    & \textit{Flux.1} & \underline{0.904} & \underline{0.945} & 0.899 & \textbf{0.659} & \underline{0.852} \\
\midrule[0.5pt]
CustFilmer (Ours)      & \textit{OmniGen2} & \textbf{0.905} & \textbf{0.961} & \underline{0.914} & \underline{0.657} & \textbf{0.858} \\
\bottomrule
\end{tabular}
}
\vspace{-5pt}
\label{tab:main}
\end{table*}

\subsubsection{Prompt Tuning for Shot-type Control.}
Cinematic storytelling requires diverse visual perspectives. To this end, we introduce a set of shot-type embeddings learned via parameter-efficient prompt tuning~\cite{lester2021power} on the Condensed Movie Dataset (CMD)~\cite{bain2020condensed}. This design enables general-purpose UMLLMs to capture the compositional priors required for cinematic keyframe generation without retraining the entire model. Let $\mathcal{K}$ denote the shot-type vocabulary, and let $k_t \in \mathcal{K}$ represent the shot-type token assigned to script $s_t$.
Given the hidden state $h_t$ at each DiT iteration, we prepend a learnable shot-type embedding $E_{\text{shot}}(k_t) \in \mathbb{R}^{d \times N}$, where $d$ is the embedding dimension and $N$ denotes the number of tokens, to construct a shot-aware representation: $h_t' = [\, E_{\text{shot}}(k_t)\,;\,h_t\,]$. The sequence $h_t'$ is then used to condition the DiT decoder: $I_t = \mathrm{DiT}(h_t', z_t)$.
This hidden-state prefixing enforces shot-specific composition without re-encoding by autoregressive model, leading to outputs that are visually diverse.

\subsubsection{Multi-subject Story Customization.}
Prior story generation methods~\cite{zhou2024storydiffusion, wang2025characonsist} are largely restricted to single-character settings, which limits their ability to model multi-subject interactions. In contrast, CustFilmer incorporates multiple reference images within visual reference retrieval, enabling multi-character story customization and substantially enhancing narrative flexibility.

\subsection{TI2V Expansion}
As a downstream step, the generated keyframes can be expanded into videos using off-the-shelf TI2V models. Rather than relying on the last generated frame (known to cause error accumulation in long-horizon generation), we treat our consistent keyframes from Stage~2 as the anchors for video extension. Given the script prompts from Stage~1 and the keyframes, we employ a TI2V model (e.g., Wan~\cite{wan2025wan}) to produce short clips conditioned on each keyframe. These clips are then concatenated into longer videos.

\begin{figure*}[!ht]
    \centering
    \includegraphics[width=.92\linewidth]{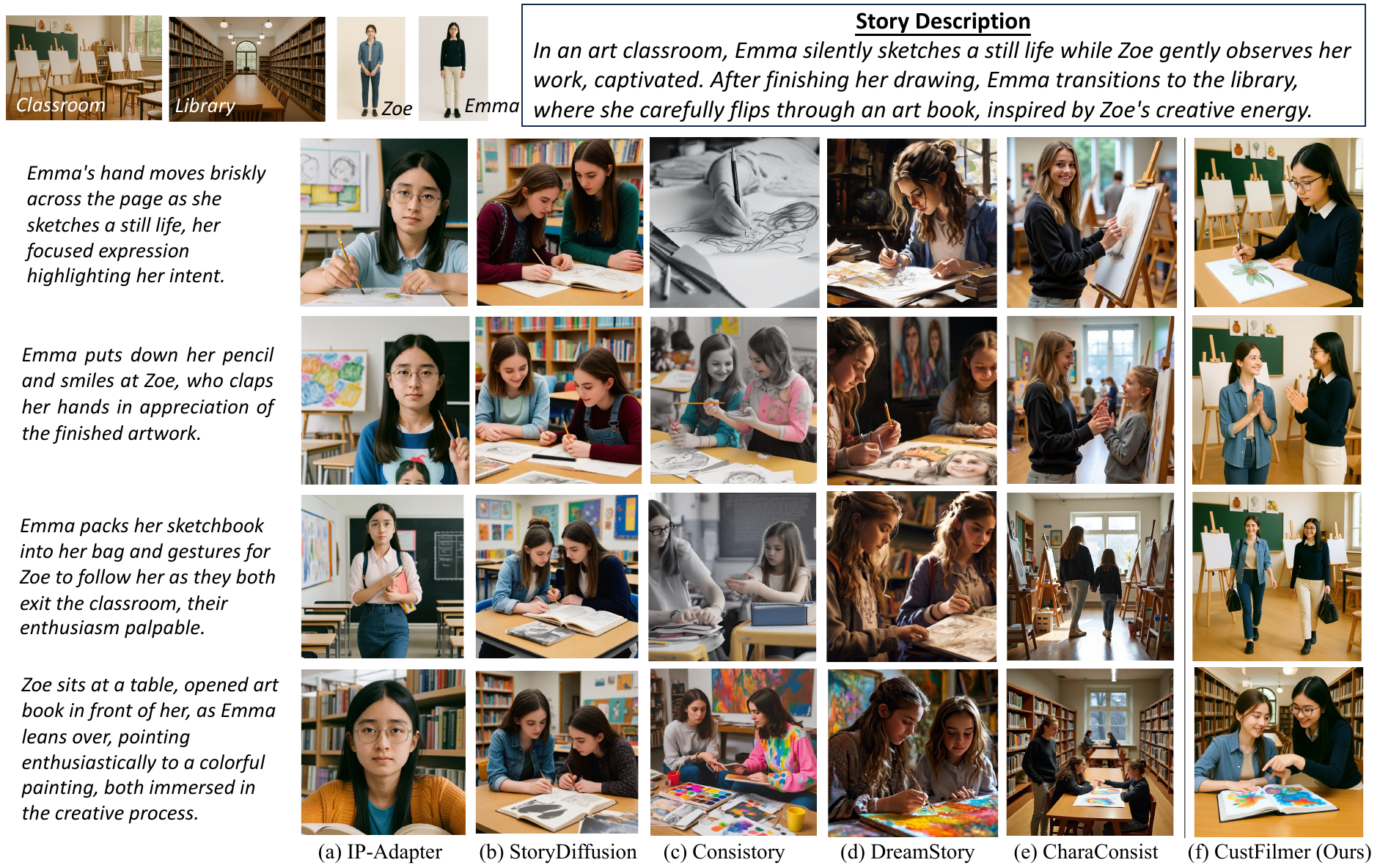}
    \caption{Qualitative Comparison between (a) IP-Adapter~\cite{ye2023ip}, (b) StoryDiffusion~\cite{zhou2024storydiffusion}, (c) Consistory~\cite{tewel2024training}, (d) DreamStory~\cite{he2025dreamstory}, (e) CharaConsist~\cite{wang2025characonsist} and (f) CustFilmer (Ours) on M$^2$SB}
    \label{fig:qc2}
    \vspace{-5pt}
\end{figure*}

\section{Experiments}
\label{sec:exp}
To comprehensively evaluate our approach, we present the experimental setup, results, and analysis.

\subsection{Experimental Setup}

\noindent\textbf{Implementation Details.}
We restructure OmniGen2~\cite{wu2025omnigen2} as the backbone. 
For CustFilmer, we optimize the shot-type prompt for 4,000 iterations using 4 NVIDIA H100 GPUs, and perform inference on a single NVIDIA 5090 GPU. 
We set $\alpha = 0.75$, 
$d=2048$ and $N=30$ in NKP.

\noindent\textbf{Datasets.} 
Existing datasets (e.g., DS500~\cite{tewel2024training}, ConsiStory+~\cite{liu2025one}) for story generation are limited to text-only inputs. To systematically evaluate the performance of our method, we propose the MSB and M$^2$SB dataset for evaluating story customization. 
MSB and M$^2$SB consist of 100 stories, each containing 8 scripts, resulting in 800 scripts. In each script, we provide a text prompt, reference images, and a shot-type annotation. To train the shot-type prompt in CustFilmer, we collect images from Condensed Movie Dataset (CMD)~\cite{bain2020condensed}. The details of the pipeline regarding MSB, M$^2$SB, and the dataset for shot-type prompt tuning are provided in Sec.~\ref{sec:msb} and~\ref{sec:rst} in the Supplementary.

\begin{table}[!t]
\centering
\caption{Comparison on text alignment/ image quality.}
\vspace{.5em}
\resizebox{1\columnwidth}{!}{
\begin{tabular}{lcccc}
\toprule
Method 
  & CLIP-T($\uparrow$) & IAS($\uparrow$) & IQS($\uparrow$) & STA($\uparrow$) \\
\midrule
IP-Adapter (Arxiv'23)       & 0.216 & 0.449 & 0.396 & 0.315 \\
Consistory (SIGGRAPH'24)     & \textbf{0.303} & 0.431 & 0.385 & \underline{0.406} \\
StoryDiffusion (NeurIPS'24) & 0.264 & \underline{0.450} & 0.414 & 0.290 \\
DreamStory (TPAMI'25)      & 0.276 & 0.438 & 0.411 & 0.280 \\
CharaConsist (ICCV'25)     & 0.265 & 0.448 & \underline{0.415} & 0.247 \\
\midrule[0.3pt]
CustFilmer (Ours)           & \underline{0.285} & \textbf{0.450} & \textbf{0.423} & \textbf{0.418} \\
\bottomrule
\end{tabular}
}
\label{tab:main2}
\vspace{-5pt}
\end{table}

\noindent\textbf{Comparison Method.}
We compare our method with both training-free and identity-reference approaches. For the training-free category, we select StoryDiffusion~\cite{zhou2024storydiffusion}, ConsiStory~\cite{tewel2024training}, DreamStory~\cite{he2025dreamstory}, and CharaConsist~\cite{wang2025characonsist}, which generate stories purely from textual descriptions without visual references. These baselines allow us to evaluate the benefit of multimodal conditioning. 

For the identity-reference category, we adopt IP-Adapter~\cite{ye2023ip}. Following~\cite{wang2025characonsist}, we crop the face region of the reference image using RetinaFace~\cite{deng2020retinaface} before feeding it into IP-Adapter. For models that only accept a single text input, we use captions of both characters and backgrounds as inputs. We aim to demonstrate that even under a more flexible, multimodal setup, our method can still maintain a level of consistency comparable to previous models.

\begin{table}[!t]
\centering
\caption{Comparison on multiple subjects story customization.}
\resizebox{.95\columnwidth}{!}{
\begin{tabular}{lccc}
\toprule
Method & CLIP-T($\uparrow$) & IQS($\uparrow$) & IAS($\uparrow$) \\
\midrule
IP-Adapter (Arxiv'23)     
  & 0.249  & 0.487 & 0.392\\
Consistory (SIGGRAPH'24)     
  & \textbf{0.302} & 0.463 & 0.382\\
StoryDiffusion (NeurIPS'24)  
  & 0.253 & \underline{0.495} & 0.415 \\
CharaConsist (ICCV'25)   
  & 0.263 & 0.492 & 0.424\\
DreamStory (TPAMI'25) &0.265&0.491&\textbf{0.440}\\
\midrule
CustFilmer (Ours) 
  & \underline{0.272} & \textbf{0.496} & \underline{0.439} \\
\bottomrule
\end{tabular}
}
\label{tab:main_mul}
\vspace{-10pt}
\end{table}

\noindent\textbf{Evaluation Metrics.}
Following previous works~\cite{wang2025characonsist}, we employ several metrics to evaluate the performance of our multimodal story generation model. CLIP-T measures the alignment between textual descriptions and generated images, while CLIP-I measures the pairwise CLIP-based image similarity. To assess consistency more precisely, we compute CLIP-I-fg and CLIP-I-bg, which evaluate the image similarity of foreground and background. We use Dinov2~\cite{oquab2023dinov2} and Segment Anything Model~\cite{kirillov2023segment} to split the foreground and background, and employ Alpha-CLIP~\cite{sun2024alpha} to compute the similarity in the mask area.

To evaluate the performance of story customization, we define inter-consistency and intra-consistency. Inter-consistency is defined as the similarity between reference images and generated images, while intra-consistency measures the similarity among generated images. In addition, we adopt Identity Similarity (ID-SIM) to assess character consistency across frames. Specifically, we use RetinaFace~\cite{deng2020retinaface} to detect facial regions, extract embeddings with FaceNet~\cite{schroff2015facenet}, and compute the similarity between faces. We also employ IQS and IAS~\cite{wu2023q} to evaluate the quality and aesthetics of generated images. We further adopt the Shot-Type Alignment (STA) score with a shot-type classifier~\cite{xie2025shot} to measure the accuracy of shot-type control.

\begin{figure}[!t]
    \centering
    \includegraphics[width=\linewidth]{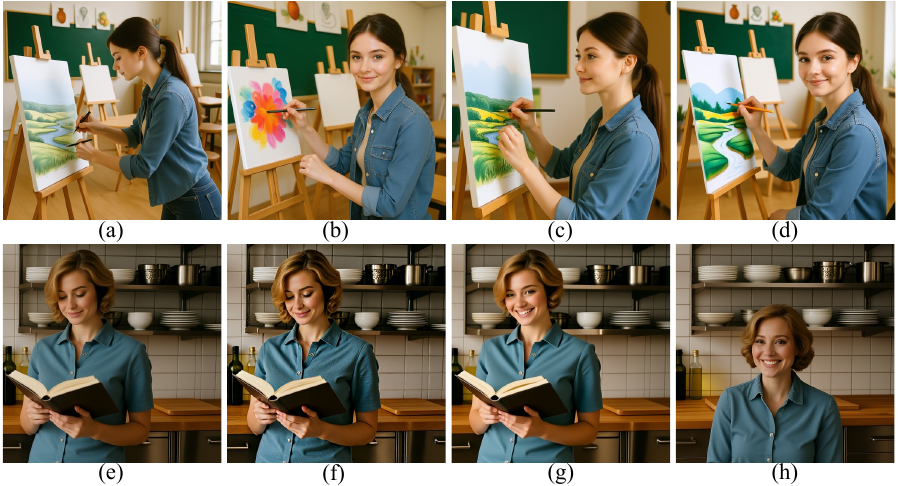}
    \caption{Top: Ablation on consistency components (a) Previous frame, (b) w/o TPC\&Retrieval, (c) +Retrieval, (d) +TPC\&Retrieval. Bottom: Ablation on $\alpha$, (e) Previous frame, (f) $\alpha=1$, (g) $\alpha=0.25$, (h) $\alpha=0.125$.}
    \label{fig:ablation_alpha}
    \vspace{-5pt}
\end{figure}

\subsection{Experimental Results}

\noindent\textbf{Quantitative Results.}
We compare CustFilmer with IP-Adapter, StoryDiffusion, ConsiStory, and CharaConsist in Tab.~\ref{tab:main} and Tab.~\ref{tab:main2} on MSB. 
As can be seen in the tables, CustFilmer 
performs the best on overall consistency, benefiting from image references for stronger inter-consistency. 
For intra-consistency, its foreground scores are lower due to richer action variations from following textual descriptions, while CharaConsist attains higher background scores by generating nearly identical scenes. This reflects a trade-off between consistency and diversity. 

In Tab.~\ref{tab:main2}, our method achieves slightly lower CLIP-T scores than ConsiStory. 
{\color{black} We attribute it to the use of different backbones. 
DreamStory is built upon Stable Diffusion, which employs CLIP text encoders during training, leading to strong alignment with the metrics stemming from CLIPs. This phenomenon is also observed in CharaConsist~\cite{wang2025characonsist}.
On multi-subject customization (Tab.~\ref{tab:main_mul}), CustFilmer achieves competitive results on M$^2$SB.

In addition, we extend CustFilmer with TI2V, and compare it with a video generation model, Wan2.2~\cite{wan2025wan}. The comparisons are provided in Sec.~\ref{sec:wan2.2} in the Supplementary.

\begin{table}[!t]
\centering
\begin{minipage}{0.48\linewidth}
    \centering
    \caption{Ablation on TPC and Retrieval on MSB.}
    \vspace{.5em}
    \resizebox{\linewidth}{!}{
    \begin{tabular}{ccc}
        \toprule
        TPC & Retrieval & Avg-Consistency \\
        \midrule
        \xmark & \xmark & 0.854 \\
        \cmark & \xmark & 0.855 \\
        \xmark & \cmark & \underline{0.856} \\
        \cmark & \cmark & \textbf{0.858} \\
        \bottomrule
    \end{tabular}}
    \label{tab:strategy_comparison}
\end{minipage}
\hfill
\begin{minipage}{0.48\linewidth}
    \centering
    \caption{Ablation on different $\alpha$ on MSB.}
    \vspace{.5em}
    \resizebox{\linewidth}{!}{
    \begin{tabular}{ccc}
        \toprule
        $\alpha$ & CLIP-T & Avg-Consistency \\
        \midrule
        0.125 & \textbf{0.2894} & 0.850 \\
        0.25  & \underline{0.2885} & 0.854 \\
        0.50  & 0.2863  & 0.857 \\
        0.75  & 0.2852  & \underline{0.858} \\
        1.00  & 0.2841  & \textbf{0.860} \\
        \bottomrule
    \end{tabular}}
    \label{tab:num_ref}
\end{minipage}
\vspace{-10pt}
\end{table}

\noindent\textbf{Qualitative Results.} 
For the qualitative comparison, we present the results of CustFilmer and other methods on single-subject story customization in Fig.~\ref{fig:qc1} and on multi-subject story customization in Fig.~\ref{fig:qc2}.
As seen in Fig.~\ref{fig:qc1} and Fig.~\ref{fig:qc2}, existing methods struggle to jointly leverage reference images and text, often leading to inconsistencies in backgrounds and character clothing, as in IP-Adapter, StoryDiffusion, and ConsiStory. CharaConsist, while maintaining stable backgrounds, produces scenes that appear overly static. In contrast, CustFilmer generates coherent sequences that preserve character identity while ensuring background consistency. Notably, our method can produce visually richer results by following shot types.


\subsection{Analysis}

\noindent\textbf{Ablation on Consistency Components.} To demonstrate the effectiveness of TPC and Visual Reference Retrieval, we show quantitative results on MSB in Tab.~\ref{tab:strategy_comparison}. Both components improve consistency, and their combination achieves the best performance. Qualitative examples in Fig.~\ref{fig:ablation_alpha} (top row) show that CustFilmer can preserve objects across frames (e.g., the evolving painting) and maintain story flow. 

\noindent\textbf{Ablation on Different Consistent Ratios ($\boldsymbol{\alpha}$).}
In Tab.~\ref{tab:num_ref}, we ablate different consistency ratios, and the corresponding qualitative comparisons are shown in Fig.~\ref{fig:ablation_alpha} (bottom row). We observe that increasing $\alpha$ leads to stronger consistency across frames, but at the cost of reduced visual diversity. This is because allocating more tokens to reference images reduces the proportion of text tokens, making the generated results less aligned with textual descriptions and thereby lowering the CLIP-T score. To balance consistency and diversity, we set $\alpha = 0.75$ in our experiments. We provide more ablation study in Sec.~\ref{more_ablation} in the Supplementary.

\section{Conclusion}
\label{sec:con}
We present VstoryGen, a multimodal framework for story generation. Our method integrates multimodal scripts to generate long visual sequences with both character and scene consistency. To achieve this, we introduced CustFilmer, which leverages the power of the multimodal generation model and employs Text Prompt Consolidation, Visual Reference Bank and Retrieval, and shot-type–aware prompt tuning to produce coherent and cinematic keyframes. We further developed MSB and M$^2$SB, two benchmarks that evaluate story generation from consistency, alignment, and shot-type perspectives. Experimental results show that CustFilmer outperforms existing methods in maintaining consistency and enabling controllable story generation. We believe this work provides a step toward more flexible and reliable multimodal story generation, with potential applications in filmmaking and storytelling.

{
    \small
    \bibliographystyle{ieeenat_fullname}
    \bibliography{main}
}

\clearpage
\setcounter{page}{1}
\maketitlesupplementary

\begin{figure*}[!t]
    \centering
    \includegraphics[width=0.95\linewidth]{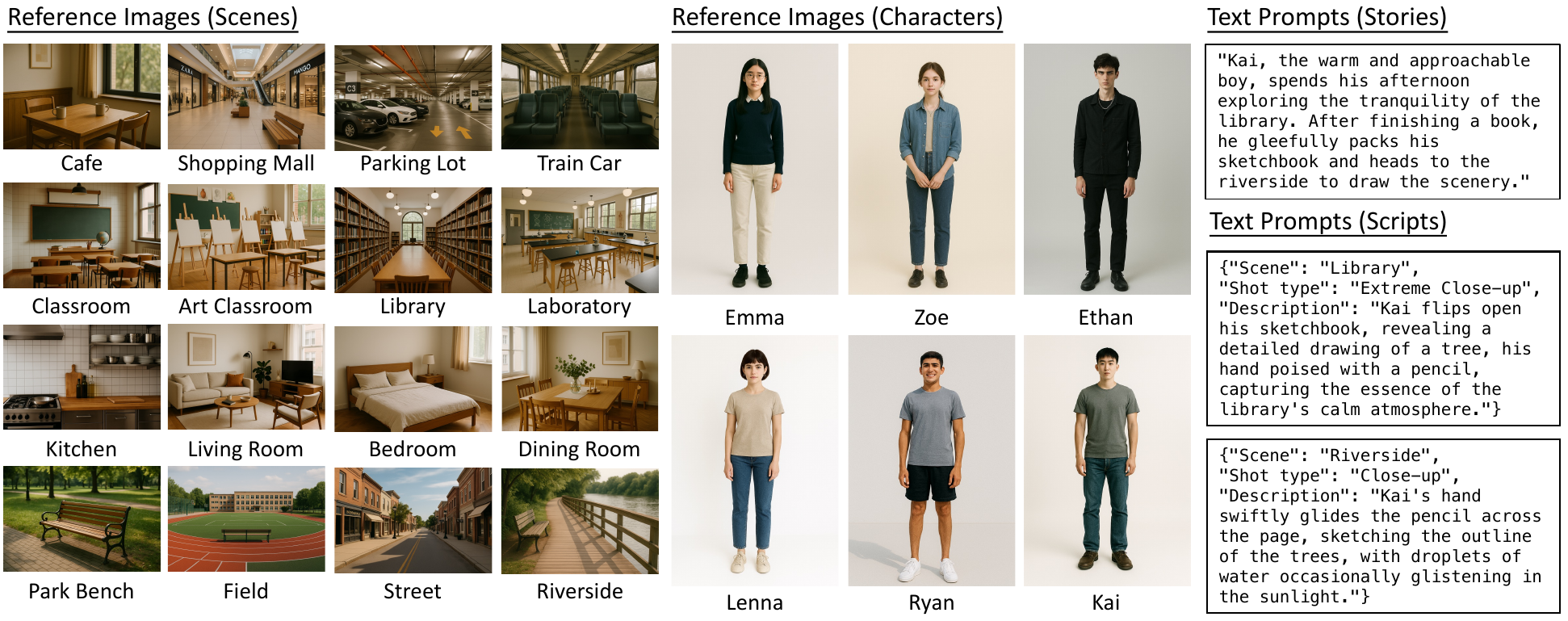}
    \caption{Overview of Multimodal Story Benchmark (MSB)}
    \label{fig:msb}
\end{figure*}

\section{Details of MSB \& M$^2$SB}
Since existing storytelling datasets~\cite{tewel2024training} primarily provide story descriptions as text-only inputs, they are insufficient for evaluating story customization, which requires multimodal conditioning such as specified characters, backgrounds, and desired shot types. To address this limitation, we introduce two new datasets, MSB and M$^2$SB, designed specifically for benchmarking multimodal story customization, as shown in Fig.~\ref{fig:msb}.

To construct this dataset, we first employ GPT-4o to generate a diverse set of story outlines along with corresponding candidate character and scene images.
We then prompt GPT-4o to produce detailed multimodal story scripts for each keyframe, including the full script prompt, reference character and background mentions, and shot-type annotations. In total, we generate 100 stories, each containing 8 multimodal scripts, resulting in 800 scripts overall. These datasets enable systematic evaluation of customization, consistency, and controllability in story visualization models. ~\label{sec:msb}
\section{Dataset for Shot-type Prompt Tuning}~\label{sec:rst}
In this section, we describe the pipeline used to construct the dataset for tuning the shot-type prompt. The dataset is built through the following five steps:
\begin{itemize}
    \item \textbf{Video collection} We begin by collecting video data from the Condensed Movie Dataset (CMD)~\cite{bain2020condensed}.
    \item \textbf{Character tracking} We apply ByteTrack~\cite{zhang2022bytetrack} to track character trajectories across frames, enabling retrieval of the same individual across different shots and scenes.
    \item \textbf{Frame pairing and identity verification} We randomly sample two frames whose temporal distance is larger than a predefined threshold, and use CLIP~\cite{radford2021learning} to verify that both frames depict the same character, thereby avoiding trivial duplicates or copy–paste artifacts.
    \item \textbf{Shot-type annotation} We apply the shot-type classifier from~\cite{xie2025shot} to categorize the target frame into one of the canonical cinematic shot types.
    \item \textbf{Caption generation} Finally, we use Qwen2.5-VL~\cite{bai2025qwen2} to generate a caption for the target frame, which serves as the textual prompt.
\end{itemize}
The resulting dataset contains 715 example pairs, which we use for training. Example pairs from this dataset are shown in Fig.~\ref{fig:rst}.

\begin{figure*}[!t]
    \centering
    \includegraphics[width=0.85\linewidth]{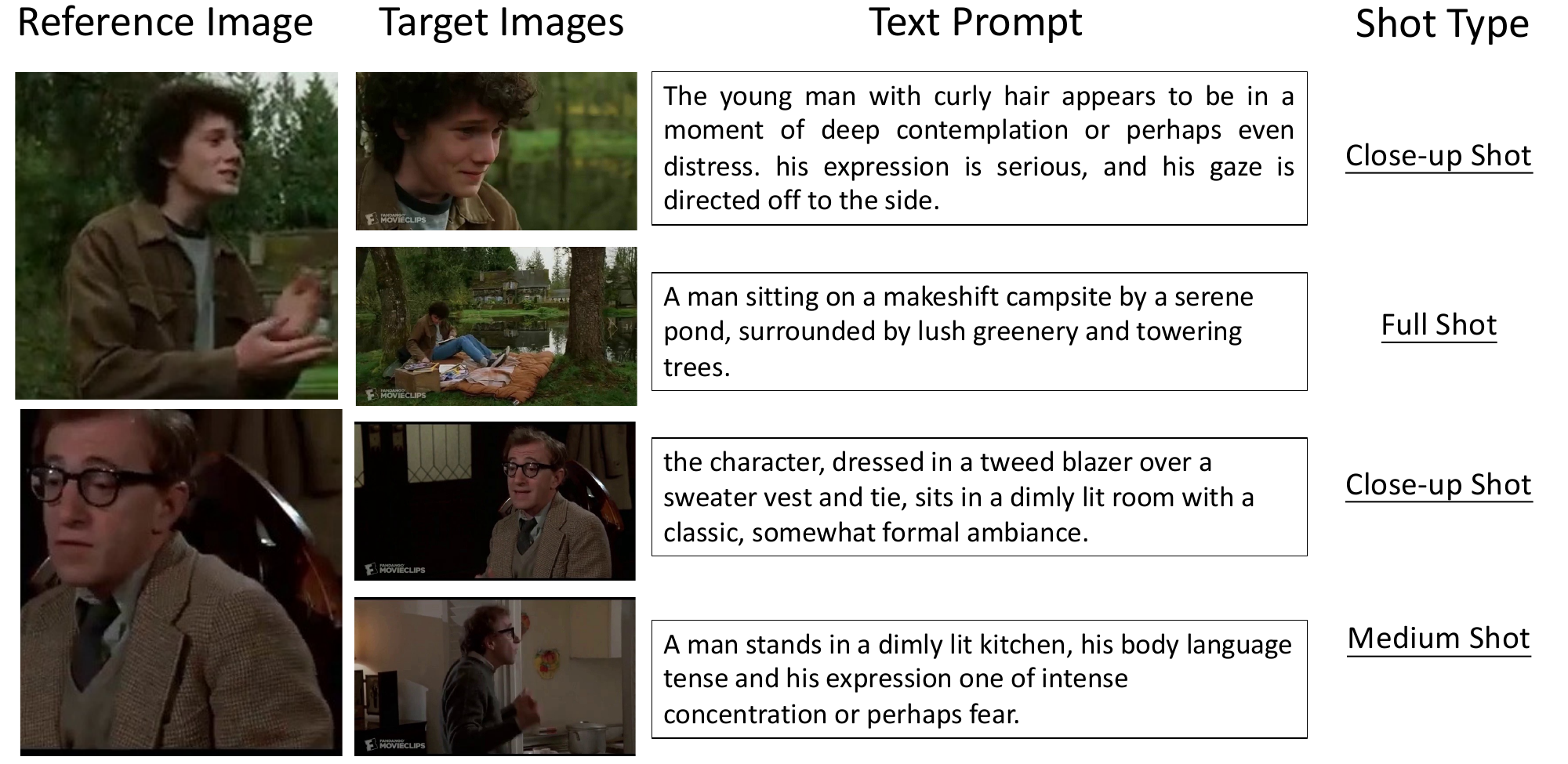}
    \caption{Dataset for shot-type control from CMD}
    \label{fig:rst}
\end{figure*}




\begin{table}[!t]
\centering
\caption{Comparison of VstoryGen and Wan2.2 on MSB}
\vspace{.5em}
\resizebox{0.65\linewidth}{!}{
\begin{tabular}{lcc}
\toprule
Method & Clip-T \\
\midrule
Wan2.2-TI2V-5B (Baseline) & 0.240 \\
VstoryGen (w/ Wan2.2)  & \textbf{0.285} \\
\bottomrule
\end{tabular}}
\label{tab:wan_comparison}
\end{table}

\begin{table}[!t]
\centering
\caption{Ablation on Visual Reference Retrieval with different numbers of reference images from previous keyframes.}
\resizebox{0.5\linewidth}{!}{
\begin{tabular}{ccc}
\toprule
$\mu$ & Clip-T & Avg-Consistency\\
\midrule
0 & 0.284 &  0.855\\
1 & \textbf{0.285} &  \textbf{0.858}\\
2 & 0.282 &  0.856\\
\bottomrule
\end{tabular}}
\label{tab:nkp_ablation}
\vspace{-5pt}
\end{table}



%

\section{More Experiments and Ablation Study}

\subsection{Comparison between Wan2.2 and VstoryGen.}~\label{sec:wan2.2}
We compare VstoryGen with the original Wan2.2~\cite{wan2025wan} model on the MSB benchmark. Since VstoryGen integrates all reference images beforehand, it produces video results that align more closely with the textual descriptions.
We compute the CLIP score between each generated video segment and its corresponding text. In Tab.~\ref{tab:wan_comparison}, our method achieves higher CLIP scores, indicating that it generates videos that are semantically closer to the given descriptions.

\subsection{Ablation on Visual Reference Retrieval.}
We ablate Visual Reference Retrieval with varying numbers of previous reference images, $\mu$. As shown in Tab.~\ref{tab:nkp_ablation}, using only the most recent reference frame achieves the best performance, while incorporating earlier frames slightly degrades results. 
We hypothesize that multiple temporally distant references introduce conflicting or less relevant visual cues. As model must align its prediction with all references, excessive conditioning may overwhelm the model and dilute its focus on the most informative frame, $I_{t-1}$. Consequently, relying on a single temporally adjacent reference allows the model to maintain a clearer and more stable temporal signal, leading to improved consistency.~\label{more_ablation}

\end{document}